# DXNN Platform: The Shedding of Biological Inefficiencies

Gene I. Sher

*Abstract*—This paper introduces a novel type of memetic algorithm based Topology and Weight Evolving Artificial Neural Network (TWEANN) system called DX Neural Network (DXNN). DXNN implements a number of interesting features, amongst which is: a simple and database friendly tuple based encoding method, a 2 phase neuroevolutionary approach aimed at removing the need for speciation due to its intrinsic population diversification effects, a new "Targeted Tuning Phase" aimed at dealing with "the curse of dimensionality", and a new Random Intensity Mutation (RIM) method that removes the need for crossover algorithms. The paper will discuss DXNN's architecture, mutation operators, and its built in feature selection method that allows for the evolved systems to expand and incorporate new sensors and actuators. I then compare DXNN to other state of the art TWEANNs on the standard double pole balancing benchmark, and demonstrate its superior ability to evolve highly compact solutions faster than its competitors. Then a set of oblation experiments is performed to demonstrate how each feature of DXNN effects its performance, followed by a set of experiments which demonstrate the platform's ability to create NN populations with exceptionally high diversity profiles. Finally, DXNN is used to evolve artificial robots in a set of two dimensional open-ended food gathering and predator-prey simulations, demonstrating the system's ability to produce ever more complex Neural Networks, and the system's applicability to the domain of robotics, artificial life, and coevolution.

*Keywords: Neural Network, Computational Intelligence, Genetic Algorithms, Artificial Life, Neuroevolution, Memetic Algorithms.*

## 1. Introduction

Neural Networks (NNs) are universal function approximators capable of modeling complex mappings between inputs and outputs. As the problem posed to the NN increases in complexity, the size of the NN must also increase, and as the size of the NN increases, the difficulty of training that NN grows with it. Even before the NNs reach moderate complexity and size, hand setting the weights, topology, and other parameters of the NN becomes impractical. The standard training algorithms like the error back-propagation[11] get stuck too easily in the local maxima of even just moderately difficult problems, like the single pole balancing task for example. The more complex problems require not only the efficient tuning of the weights, but also an automated method of changing the topology of the neural systems to fit the problem. An efficient solution for an automated method of setting both, the topological and parametric parameters in the NN, is accomplished through the application of evolutionary algorithms (EA). By using EA to evolve weights, general parameters, and topology of the NN, highly complex problems can be solved. The NN systems capable of topology and weight evolving through EA are referred to as Topology and Weight Evolving Artificial Neural Networks, or TWEANNs. There are many approaches and algorithms when it comes to TWEANNs, but even with these highly advanced systems the "curse of dimensionality"[12] and other problems persist, pushing us to continue the search for ever more sophisticated TWEANN systems. One such new system is introduced in this paper.

The proposed NN system referred to as: DX Neural Network (DXNN), presents a number of new features and improvements over the currently published TWEANNs. DXNN allows us to further accelerate the production of topological solutions to various problems, while at the same time consistently producing much more compact NNs when compared to even the most advanced TWEANN systems [3][4]. Through a 2 phase approach to neuroevolution and with the utilization of a Targeted Tuning Phase, DXNN produces populations of ever increasing diversity and complexity. Furthermore, DXNN's simple built-in feature selection algorithm allows it to dynamically establish links to new sensors and actuators as they become available, making DXNN highly fit for problems in the field of artificial life and robotics, as will be shown in the later sections.

This paper shall be organized as follows: Section 2 will introduce the general functionality of TWEANN systems and list the modern state of the art TWEANNs with short descriptions of each. Section 3 will provide a detailed description of the DXNN platform's architecture. Section 4 will discuss DXNN platform's learning algorithm. Section 5 will discuss and compare the results of the standard double pole balancing benchmark for DXNN and other TWEANNs. Section 6 will demonstrate that DXNN produces exceptionally high population diversity due to its hybrid approach to evolution. In section 7 I shall perform feature variation and oblation experiments, demonstrating how the various features affect DXNN's performance. Section 8 will demonstrate the results of applying DXNN to an open-ended food gathering and predator-prey simulations, demonstrating interesting results where predators learn to hide behind plants while pushing them around to bait and ambush prey. Finally, section 9 will end with a summary and future plans for this system.

Note: In this document, "DXNN Platform" refers to the entire software package which builds NNs, supervises and monitors the NN population, and applies the mutational operators to the NN. Both, NN and DXNN are used interchangeably when referring to the actual Neural Networks generated by the DXNN Platform. Finally, DX in DXNN stands for Deus Ex, Latin deus, god + Latin ex, from. The goal of computational intelligence through evolution, for some of us, is the creation of true intelligence, not constrained by biology as we are. It is with this in mind that the name for this system was chosen.

## 2. Modern TWEANN algorithms:

As the problem to be solved increases in complexity, a static NN becomes too inflexible to tackle it. If the NN is static, its size and topology must be guessed and hand designed beforehand by the researcher for every problem. Since this type of knowledge is very difficult to guess at from the problem itself, and further becomes impossible when the problem reaches a non trivial level of complexity, an automated method for doing so becomes a necessity. If the problem itself changes as times goes on, then a static NN, no matter how expertly set beforehand, will not be able to deal with the changing problem. Furthermore, to mitigate the limitations of local search algorithms, an advanced system must be able to pull itself out of, or avoid altogether the local maxima that riddle the fitness landscapes of complex problems. TWEANN systems are the solution to such and other problems.

In TWEANN systems a seed population of NNs is created, each with some initial minimal or random topology of interconnected neurons. Each NN within the population then attempts to solve the problem at hand, and based on its performance is assigned some fitness value. The individuals within the population are then compared with one another based on their fitness and the use of a selection algorithm. Afterwards, the most fit of the individuals are allowed to produce offspring while the unfit individuals are discarded. The offspring of the successful NN is usually either a mutated copy of the fit NN itself, or a combination of two or more NNs through some crossover method. In this manner, through mutations or crossover, weights and topology of the NNs are changed, and through evolutionary pressure for higher fitness, new and superior NNs are generated.

Though as advanced as these new methods are, there are still problems in the field waiting to be solved. For example, as the complexity of the problem and the size of the NN grows, more and more weights, connections, and other parameters have to be set in just the right way to produce a solution. This problem is referred to as "the curse of dimensionality" and many state of the art TWEANN systems become stuck in local maxima on the fitness landscape. Another problem is that some TWEANNs begin to add more and more neurons to the network while producing only very minimal fitness gains, this leads to increasingly bloated NN solutions overtaking the population. Ironically, this bloating results in even more weights, parameters, and links that need to be set up concurrently to produce a solution, thus making it even harder, and eventually impossible [7], for the NN to improve any further and solve the problem. The DXNN Platform proposes and demonstrates algorithms to mitigate these and other setbacks.

**2.1 Existing methods:**

There are a number of existing algorithms that evolve NNs. Among such methods are the following state of the art systems: EPN[1], GNARL[2], NEAT[3], CoSyNe[4], HyperNEAT[5], EANT[6], and EANT2[7].

EPN: uses augmented back-propagation for weight optimization, and addition and removal of Neurons as topological mutation operators. EPN does not employ crossover.

GNARL: utilizes EA for both weight and topological optimization, utilizing a "temperature" parameter to determine the intensity of random mutations, a concept similar to the one used in simulated annealing [13]. Like EPN, GNARL also avoids utilizing crossover.

NEAT: uses genetic algorithms to both mutate the weights and the topology. The weights are mutated through small perturbations, and the topological mutation operators are composed of: adding links to an existing neuron, adding a new neuron, crossover, and splicing. Splicing is described as the following process: two neurons connected to each other are first disconnected and then reconnected through a newly created neuron. NEAT also employs speciation, separating the NNs based on their topologies, trying to preserve diversity and allowing for the unfit individuals to survive for a few extra generations in hopes that they have the potential for improvement.

CoSyNe: uses a cooperative co-evolution approach, where various permutations of neurons belonging to different groups are tried in combination to determine which of them work best in most combinations. Based on such criteria the final NN is then composed by combining the best and most generally fit neurons.

HyperNEAT: is an extension of NEAT. In HyperNEAT, NEAT is used to drive a substrate of Neurons. The substrate of Neurons is composed of a neural grid. Each neuron on the grid has a coordinate distributed uniformly between -1 and 1 on all axis of the substrate. Instead of using the NN to solve the problem directly, HyperNEAT uses the NN to generate weights for the Neurons on the grid, also referred to as a substrate. The weight is generated for every Neuron by letting NEAT produce a weight based on the coordinate of a given Neuron and the coordinate of the Neuron linking to the given Neuron. When the substrate is a plane, the problem becomes a 4 dimensional one with the [X1,Y1,X2,Y2] used as the input to the NN and [W] the output. Thus, on a two dimensional substrate a weight W of a neuron A with coordinates [Xa,Ya] connected from a neuron B with coordinates [Xb,Yb] is determined by feeding the NN a vector [Xa,Ya,Xb,Yb]. This type of indirect encoding has shown to produce interesting generalization capabilities. In this paper I shall refer to this type of encoding as Substrate Encoding (SE).

EANT2: separates the learning approach into two steps, exploitation through CMA-ES ("Covariance Matrix Adaptation Evolution Strategy")[8], and then exploration through standard topological mutations. It too does not utilize mating.

Neither CoSyNe nor CMA-ES evolve a topology, instead they optimize a single general topology. Thus these approaches can not be used for complex problems or open ended problems like Artificial Life. When using these two methods the topological solution for the problem must be known beforehand. Since topology is not evolved, it is not possible to generate minimal topologies for the problems either, unless such topologies are already known.

## 3. DXNN Topology and Elements:

The DXNN platform was created with scalability, robotics, and artificial life in mind. It was created to be implemented by concurrent languages like Erlang and OCaml, and utilize the full power of distributed multi-core and multi-CPU hardware. The platform was made with the ability to change the functionality of activation functions, axonal linking methods, learning methods, and other features by simply updating the configuration parameters of the independent processing elements: Neurons, Sensors, Actuators, and the Core. In DXNN each node is represented as an independent mini server/client with its own address and concurrency to all other nodes. In the following sections I will first discuss the DXNN's general architecture, and then follow by an elaboration on every element, its role, and its genotypical encoding.

### 3.1 The General Architecture:

With DXNN being written purely in the Erlang programming language, every Neuron, Sensor, Actuator, and Supervisory element called Core, are independent concurrent processes which can only communicate with each other through message passing. DXNN is composed of 2 structural levels. At the lowest level are Neurons that form the NN. At the second level is the Core which synchronizes & supervises this NN, and interfaces it with its sensors and actuators. In DXNN a Neuron can utilize any type of activation function, such as a sigmoid, Gaussian, sine... What input the NN gets and what its output is used for is determined by the supervising Core. The Core is the element which controls and deals with pre/post processing of data, I/O of the sensors and actuators, and interfacing with the OS. Core polls the sensors for data which come in vectors and passes those vectors to the appropriate input layer Neurons. The Core then gathers the processed signals from the Neurons in the output layer, post-processes these output signals to produce output vectors, packages each vector in the form appropriate for the actuator it is destined for, summons the actuator programs, and finally passes each actuator its own vector [Fig1].

**Fig. 1** The hierarchical structure of DXNN. Core and its supervised Neurons, interfacing with sensors and actuators

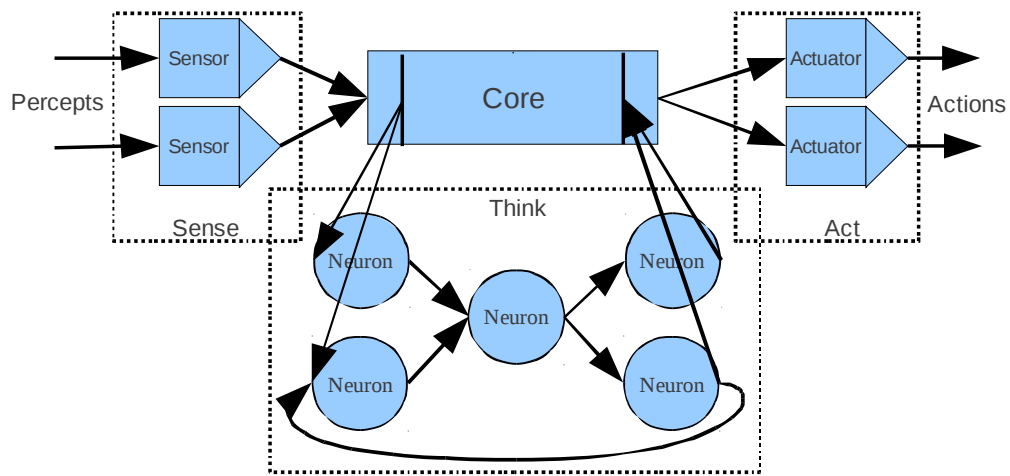

To elaborate in detail on how the DXNN Platform functions, I will first cover the functionality of the Core, Neuron elements, and the flow of information within the system, and then follow up with the discussion on DXNN Platform's learning algorithm.

## 3.2 Core:

Core is the supervisor of the entire NN, it is a program that is the interface between the actual NN and the OS/Environment/Sensors/Actuators. In its phenotypic form it is represented as a mini server/client with an Id/Address, a SensorList, an ActuatorList, a ParameterList that can further augment the DXNN's functionality, and a list of Neuron_Ids that the Core supervises. Within a genotype it is represented as a tuple: {Id, SensorList, ActuatorList, ParameterList, SupervisedNeuronIds, Generation, History}. SensorList is itself a list of tuples where each tuple is composed of a tag representing the name of a SensorProgram that the Core needs to run to get an input vector associated with a particular sense and an associated list of Neuron Ids/Addresses which should receive the signal from that SensorProgram. SensorList can be represented as follows: [{[Neuron_Id1...], InfraredSensor_Id}...{[Neuron_Id1...], ChemicalSensor_Id}]. ActuatorList is a list of tuples where each tuple is composed of a tag representing a name of an ActuatorProgram and an associated list of Neuron Ids/Addresses whose outputs are gathered, packaged and then forwarded to the ActuatorProgram. ActuatorList can be represented as follows: [{[Neuron_Id1...], LegServos_Id}...{[Neuron_Id1...], CameraTiltPanServos_Id}]. When the Neurons send signals to the Core, the Core gathers and sorts these signals for each Actuator, then calls the associated ActuatorProgram and passes it its accumulated vector. The ActuatorProgram parses the vector and executes its function, whether it be moving a virtual agent, writing a value to database, moving an actual robot by driving the servos, or even modifying some part of the NN's own topology.

The "Generation" variable is an integer that increments every time the NN the Core supervises goes through a topological mutation phase. "History" is a list composed of all the mutations applied to the NN, listed in the order they were applied. The History list is composed of the following tuple: {MutationOperator, ElementAppliedTo, Info}. Where the MutationOperator is a tag/name of the mutation operator, ElementAppliedTo is an Id, and Info is extra information, if any, and depends on the type of MutationOperator.

When connected, Core, Neurons, Sensors, and Actuators might function as in the following example: The Core might begin by going through the SensorList, calling the programs and passing the resulting vectors to the appropriate Neurons. A SensorProgram can be one that polls a camera for data and then encodes an image as a vector of length n: [Val1,Val2,Val3...Valn] where Val is a scaled floating point. This vector is then passed to the input layer Neurons for processing. At some later point, the Neurons in the output layer pass to the Core their output signals. Based on the Id/Address of the output layer Neurons, the Core chooses the appropriate ActuatorProgram, and then passes to that actuator the accumulated Vector. The ActuatorProgram can for example control the servos to move a camera [Fig2] by sending it some signal which it derives by processing the vector. An example would be a vector: [Val1,Val2], which can represent the pan and tilt signals respectively. Afterwards, the Core again polls the sensors in the SensorList, and repeats the Sense-Think-Act cycle.

**Fig. 2** DXNN whose sensor is gathering image data from camera, and whose actuator is sending pan/tilt signals to the camera's servos

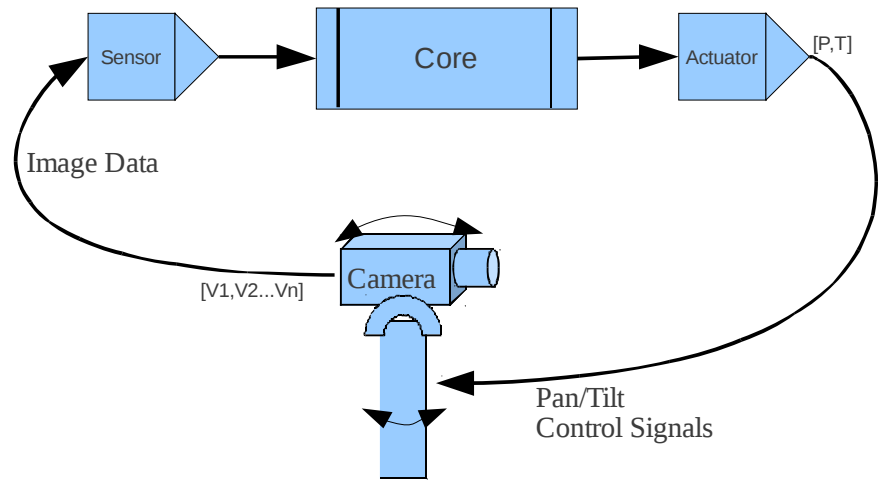

### 3.3 Neuron:

A Neuron accepts vectors as inputs and outputs a resulting vector of length 1 [Fig3]. In its phenotypic form a Neuron is a mini server/client program with an Id/Address, InputList, OutputList, ActivationFunction, LearningMethod, a WeightList, and a ParameterList which might further augment the Neuron's functionality. Within a genotype it is represented as a tuple: {Id, InputList, OutputList, ActivationFunction, LearningMethod, WeightList, ParameterList, Generation}. Neurons can have any type of Activation Function, from sigmoid to the mexican hat function. During the initial phase of generating a population of minimal NN systems, and during the topological mutation phase, a random Activation Function (AF) is chosen from a list of available AF programs represented by a list of tags. In such a list each tag is a program name that can operate on a value passed to it. Thus, as soon as a new AF program is created, the name of that program can be added to the existing list as a tag, and later during a topological mutation phase be acquired by some neuron.

Neurons also have a Learning Method (LM) which determines how to change the neuron's weights over time. A LM is a program which accepts 3 parameters, a current weight list, an input vector, and an ActivationFunction. The output of the LM is an updated weight list and an output vector. The LM can be "none" and output the same weight list it was originally passed with an output equaling to the ActivationFunction applied to the dot product of the weight list and input vector. Alternatively the LM can be "hebbian" and output an appropriate output vector and a modified weight list by applying the hebbian learning algorithm and using the ActivationFunction on the dot-product. Like the AF list, the LM list can also be easily expanded, letting future neurons to stumble upon new LMs through mutation or acquire them when initially created. Finally, all Neurons are initially created without a bias and can acquire that bias input through mutation.

**Fig. 3** A Neuron accepts input vectors, passes them with Weights and an AF to the LM function, producing an output vector of length 1:[O], and an updated list of weights

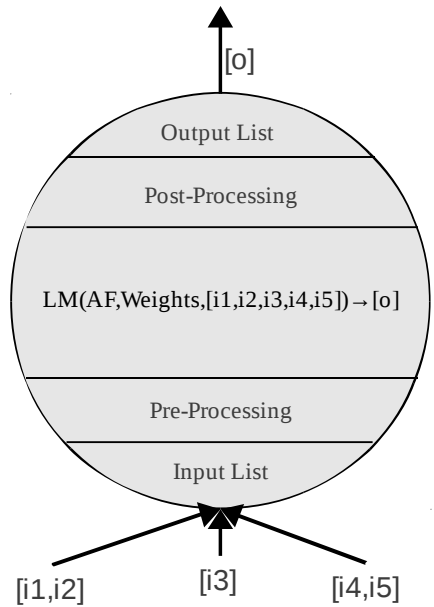

Both, Neurons and the Core, have a Generation variable. The Core's Generation is incremented every time it participates in the topological mutation phase, while the Generation variable of a Neuron is reset to that of the Core's whenever it is affected during the topological mutation phase. During the population initialization all Elements start with Generation equaling to 0. The Generation variable and the History list is essential to the selection and mutation algorithms used by DXNN, and will be elaborated upon in a later section.

**Fig. 4** The DXNN architecture, and its diagrammed 3 step Sense-Think-Act operating cycle

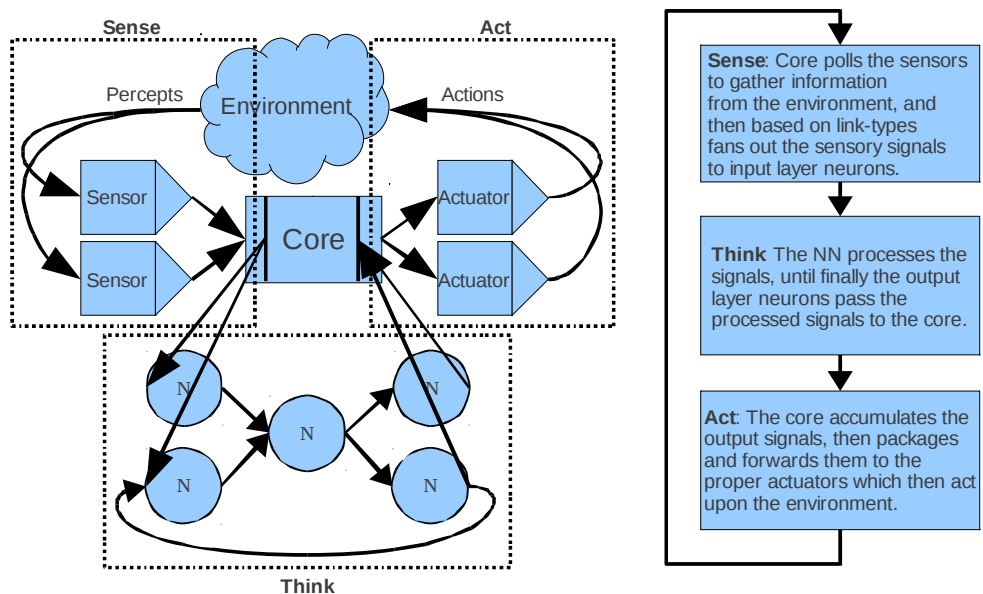

Putting the Core and the Neurons together into one system [Fig4], produces the following flow of information: The Core's list of sensors produce data vectors and distributes them to the appropriate Neurons in the input layer of the NN. The NN processes these vectors and produces an output that is passed to the Core. The Core receives this output from the output layer Neurons, packages it into vectors, and passes those vectors to their appropriate actuator programs. The actuators parse the vectors that are passed to them, and then act upon the environment. Afterwards, the Core polls the sensor programs for new data vectors once again, repeating the Sense-Think-Act cycle anew.

### 3.4 Representing DXNN Inside a Database:

The following encoding is used to represent DXNN genotype within a database:
- Population: {Population_Id, DXNN_Id_List}
  - DXNN_Id_List: [DXNN_Id1,DXNN_Id2...DXNN_Idn]
- DXNN: {DXNN_Id, Core_Id, ElementList}
  - ElementList: [ElementTuple1...ElementTupleN]
- Core Element:
{Id, SensorList, ActuatorList, ParameterList, SupervisedNeuronIds, Generation, History}
- Neuron Element:
{Id, InputList, OutputList, ActivationFunction, LearningMethod, WeightList, ParameterList, Generation}

Most of the elements within these tuples are lists themselves and are represented in a similar fashion. Since these are all nothing but lists of tuples, they are very easy to store in a relational database and traverse through. For example, to get at any Neuron one only needs a program that asks the Population element for a DXNN_Id, the proper DXNN then provides the Core_Id, and the Core_Id leads to the requested Neuron_Id. In this fashion, any mutation can be applied and any resulting topological perturbation due to the mutation can be followed and applied through these Id links. Finally, by being tuple encoded the genotype becomes human readable, easy to understand, and reasoned about. This tuple based encoding represents directly each element as a node in a directional graph, each with its in-ports and out-ports. Because this encoding gives the researcher the most direct ability to think about and visualize the system, it makes it that much easier to work with it, and expand upon it.

## 4. The DXNN algorithm:

The DXNN Platform's learning algorithm is divided into multiple stages. The "Initialization" which is executed only once to create the seed minimalistic population of DXNNs. The "Tuning Phase" in which the DXNNs interact with the environment or some problem, and undergo parametric mutation. The "Selection Stage" during which some DXNNs are put into the fit (valid) group and others into unfit (invalid) group, letting only the valid DXNNs create offspring and themselves survive into the next generation. Finally followed by the "Topological Mutation Phase" during which mutational operators are applied to the valid DXNNs, affecting topology of the Neural Network, and the various non weight parameters of the system in general. When all the stages complete, the DXNNs and their offspring (mutated versions of the valid DXNNs) are released back into the environment if the experiment is Artificial Life (ALife), or applied again to the problem during the non ALife experiments. The two main phases in this setup that designate DXNN as belonging to the field of memetic computing is the *Tuning Phase*, which is a local search implemented using a modified stochastic hill climbing, and *Topological Mutation Phase*, which is a population based global search. The following sections elaborate on each stage of the algorithm.

### 4.1 Initialization Stage:

During the initialization, every element created has its Generation set to 0. Initially a seed population of size X is created. Each DXNN in the population starts with a minimal network, where the minimal starting topology depends on the total number of Sensors and Actuators the researcher decides to start the system with. If the DXNN is set to start with only 1 Sensor and 1 Actuator with vector length of 1, then the DXNN starts with a single Core containing a single Neuron. For example, if the output is a vector of length 1 like in the Double Pole Balancing (DPB) control problem, the Core contains a single Neuron. If on the other hand the DXNN is initiated with N number of Sensors and and K number of actuators, the seed Cores will contain 2 layers of fully interconnected Neurons. The first layer which contains S Neurons, and the second A1+...Ak Neurons. Where S is the total number of Sensors, and A is the size of the vector that is destined for each Actuator. It is customary for the DXNNs to be initialized with a single Sensor and a single Actuator, letting the DXNNs discover any other axillary Sensors and Actuators through topological evolution.

Furthermore, the link from a Core to a Neuron can be of 3 types listed below:
1. Single-type link, in which the Core sends the Neuron a single value from one of its Sensors.
2. Block-type link, in which the Core sends the Neuron an entire vector that is output by one of the Sensors.
3. All-type link, in which the Core sends the Neuron a concatenated list of vectors from all the Sensors in its SensorList.

All this information is kept in the Core, the Neuron neither knows what type nor originally from which sensor the signal is coming. Each neuron only keeps track of the list of nodes it is connected from and the vector lengths coming from those nodes. Thus, to the Neuron all 3 of the previous link-types look exactly the same in its InputList, represented by a simple tuple {From_Id, Vector_Length}. The Vector_Length variable might of course be different for each of those connections.

The different link-types add to the flexibility of the system and allow the Neurons to evolve a connection where they can concentrate on processing a single value or an entire vector coming from a Sensor, depending on the problem's need. I think this improves the general diversity of the population, allows for greater compactness to be evolved, and also improves the NN's ability to move through the fitness landscape. Since it is never known ahead of time what sensory values are needed and how they need to be processed to produce a proper output, different types of links should be allowed.

For example, a Core is routing to the Neurons a vector of length 100 from one of its Sensors. Assume that a solution requires that a Neuron needs to concentrate on the 53rd value in the vector and pass it through a cosine activation function. To do this, the Neuron would need to evolve weights equaling to 0 for all other 99 values in the vector. This is a difficult task since zeroing each weight will take multiple attempts, and during random weight perturbations zeroing one weight might un-zero another. On the other hand evolving a single link-type to that Sensor has a 1/100 chance of being connected to the 53rd value, a much better chance. Now assume that a solution requires for a neuron to have a connection to all of the 100 values in the vector. That is almost impossible to achieve, and would require at least 100 topological mutations if only a single link-type is used, but has a 1/3 chance of occurrence if we have block, all, and single type links at our disposal. Thus the use of Link-Types allows the system to more readily deal with the different and wide ranging lengths of signal vectors coming from the Sensors, having a better chance of establishing a proper connection needed by the problem in question.

**Fig. 5** A DXNN that uses a substrate encoded based architecture. In this figure the Sensors pass the signals to the Core, which packages and passes the signals to the Substrate, which produces output signals and passes those to the Core which postprocesses them and passes them onwards to the Actuators. The Substrate uses the NN to set the weights of the substrate embedded neurodes

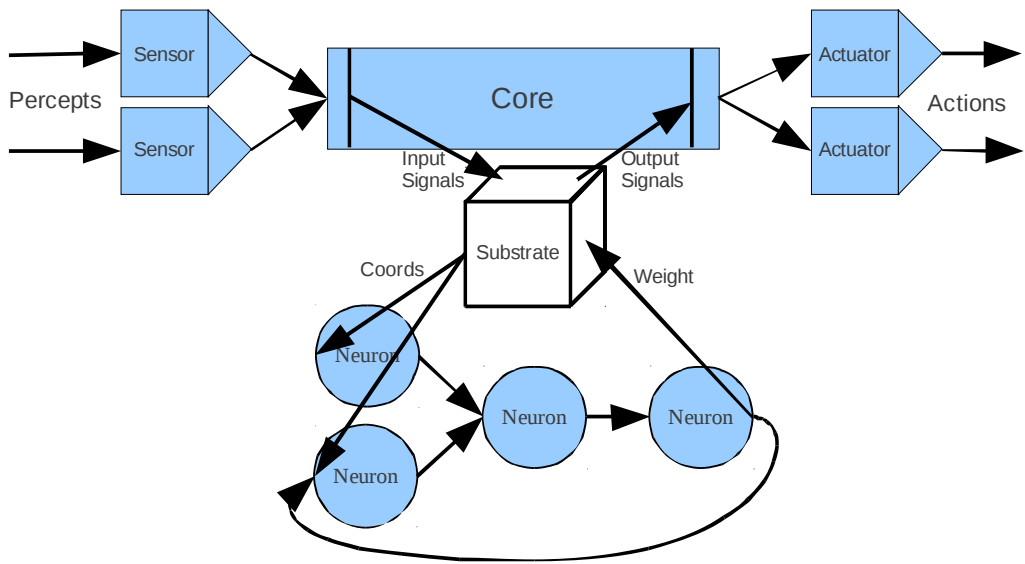

In a population, the Cores themselves can also be of different types: Type = "neural", and Type = "substrate". The "neural" type Core is one that supervises a standard recursive Neural Network. The "substrate" type Cores use their supervised NNs to drive a neural substrate, an encoding that was popularized by HyperNEAT[5]. In such Cores the sensory vector is routed to the substrate and the output vector that comes from the substrate is parsed and routed to the actuators. The supervised NN itself is polled to produce the weights for the embedded neurodes in the substrate. The type of substrates can further differ in density, and dimensionality. A digram of the DXNN architecture that utilizes a substrate is shown in Fig5. This paper will concentrate on the neural encoded version of DXNN.

## 4.2 Tuning Phase:

The first step that must be taken is to construct the phenotypic representation of the Core and its Neurons for every DXNN using the genotype stored as a list of tuples within the database. The database is scanned for the {DXNN_Id, Core_Id, ElementList} tuples, each tuple has its own Id to identify each separate DXNN. The ElementList containing the Core and Neuron tuples is then analyzed and depending on whether the correlated tuple represents a Core or a Neuron, a proper independent mini server/client is summoned for each, with the parameters and links specified by the data in that tuple. The Core then composes a list of New Generation Neurons (NGN) using the following steps:
1. Neuron Ids are sorted based on the neuron's generation, most recent(highest) to least recent(lowest).
2. Ids belonging to the 3 most recent generations are extracted, and designated as CurGenIds.
3. Square root of the total number of remaining Ids is calculated, and then this number of Ids is extracted from the remaining Id list, starting from the most recent side. We designate this Id list: RecentGenIds.
4. NGN = concatenate(CurGenIds,RecentGenIds)

After NGN is composed, a variable MaxMistakes is created and set to BaseMaxMistakes + sqrt(TotWeights from NGNs) rounded to the nearest integer. The BaseMaxMistakes variable is set by the researcher. Finally, a variable by the name AttemptCounter is created and set to 1.

The reason for the creation of the NGN list is due to the weight perturbations being applied only to the subset of these new or recently modified Neurons, a method I refer to as "Targeted Tuning". The reason to only apply perturbations to the NGNs is because evolution in the natural world works primarily through complexification and elaboration, there is no time to re-perturb all the neurons in the network after some minor topological or other type of addition to the system. As NNs grow in size it becomes harder and harder to set all the weights and parameters of all the Neurons at the same time to such values that produces a fit individual. A system composed of thousands of neurons might have millions of parameters in it. The odds of finding proper values for them all by perturbing random weights in random Neurons throughout the entire system after some minor topological mutation, all at the same time, is slim to none. The problem only becomes more intractable as the number of Neurons continues to grow. By concentrating on tuning only the newly created or newly topologically/structurally augmented Neurons and making them work with an already existing Neural Network, we make the problem much more tractable. Indeed in many respects it is how complexification and elaboration works in the biological NNs. In our organic brains the relatively recent evolutionary addition of the Neocortex was *not* done through some refurbishing of an older NN structure, but through a completely new addition of neural tissue covering and working with the more primordial parts. The Neocortex works concurrently with the older regions, contributing and when possible overwriting the signals coming from our more ancient neural structures.

During the Tuning Phase each DXNN tries to solve the problem. Afterwards, the DXNNs receive fitness scores based on their performance in that problem. After being scored each NN temporarily backs up its parameters. Every neuron in the NGN list has a probability of 1/sqrt(Tot_NGNs) of being chosen for weight perturbation. The Core sends these randomly chosen Neurons a request to perturb some of their weights. Each chosen Neuron when receiving such a request then perturbs its own weights. The total number of weights to be perturbed is chosen randomly by every Neuron itself. The number of weights chosen for perturbation by each neuron is a random value between 1 and square root of total number of weights in that Neuron. The perturbation value is chosen with uniform distribution to be between -(WeightLimit/2) and (WeightLimit/2), where the WeightLimit is set to Pi. By randomly selecting the total number of Neurons, the total number of weights to perturb, and using such a wide range for the perturbation intensity, we can achieve a very wide range of parametric perturbation. Sometimes the DXNN might have only a single weight in a single Neuron perturbed slightly, while at other times it might have multiple Neurons with multiple weights perturbed to a great degree. This allows the DXNN platform to make small intensity perturbations to fine tune the parameters, but also sometimes very large intensity (number of Neurons and weights) perturbations to allow DXNN to jump over or out of local maxima, an impossibility when using only small perturbations applied to a small number of Neurons. This high mutation variability method is referred to in the DXNN platform as a Random Intensity Mutation (RIM). The range of mutation intensities grows as the square root of the total number of NGNs, as it logically should since the greater the number of new or recently augmented Neurons in the DXNN, the greater the number of perturbations that needs to be applied to make a significant affect on the information processing capabilities of the system. At the same time the number of Neurons and Weights affected during perturbation is limited only to the newly/recently topologically augmented elements, so that the system can try to adjust the newly added structures and those elements that are directly affected by them through new connections, to work and positively contribute to an already existing neural system.

After all the weight perturbations have been applied within the DXNN, it attempts to solve the problem again. If the new fitness achieved by the DXNN is greater than the previous fitness, then the new weights overwrite the old backed up weights, the AttemptCounter is reset to 1, and a new set of weight perturbations is applied to the DXNN. Alternatively, if the new fitness is not greater than the previous fitness, then the old weights are restored, the AttemptCounter is incremented, and another set of weight perturbations is applied to the individual.

When the DXNN's AttemptCounter reaches the value of MaxMistakes, implying that a MaxMistakes number of unsuccessful RIMs have been applied in sequence without a single one producing an increase in fitness, the DXNN with its final best fitness and the correlated weights is backed up to the database through the conversion back to a list of tuples followed by a shut down of the DXNN itself. Utilizing the AttemptCounter and MaxMistakes strategy allows us, to some degree at least, test each topology with varying weights and thus let each DXNN after the tuning phase to represent roughly the best fitness that its topology can achieve. In this way there is no need to forcefully and artificially speciate and protect the various topologies since each DXNN represents roughly the highest potential that its topology can reach in a reasonable amount of time after the tuning phase completes. This allows us to judge each DXNN based purely on its fitness. If one increases the BaseMaxMistakes parameter, on the average each DXNN will have more testing done on it with regard to weight perturbations, thus testing the particular topology more thoroughly before giving it the final fitness score. On the other hand the MaxMistakes parameter itself grows in proportion to the square root of the total sum of NGN weights that should be tunned, since the greater the number of new weights that need to be tuned, the more attempts it would take to properly test the various permutations.

### 4.3 Selection Stage:

There are many TWEANNs that implement speciation during selection. Speciation is used to promote diversity and protect unfit individuals who in current generation do not posses enough fitness to get a chance of producing offspring or mutating and achieving better results in the future. Promoters of speciation algorithms state that new ideas need time to develop and speciation protects such innovations. Though I agree with the sentiment of giving ideas time to develop, I must point to [7] in which it was shown that such artificial and forced speciation and protection of unfit organisms can easily lead to neural bloating. DXNN platform does not implement forced speciation, instead it tests its individuals during the Tuning Phase and utilizes natural selection that also takes into account the complexity of each DXNN during the Selection Stage. In my system, as in the natural world, smaller organisms require less energy and material to reproduce than their larger counterparts. As an example, for the same amount of material and energy that is required for a human to produce and raise an offspring, millions of ants can produce and raise offspring. When calculating who survives and how many offspring to allocate to each survivor, the DXNN platform takes complexity into account instead of blindly and artificially defending the unfit and insufficiently tested Neural Networks. In a way, it can also be thought that every DXNN topology represents a specie in its own right, and the tuning phase concisely tests out the different parametric permutations of that particular specie, same topologies with different weights. I believe that speciation and niching should be done not forcefully from the outside by the researcher, but by the artificial organisms themselves within the artificial environments they inhabit, if their environments/problems allow for such a feat. When the organisms find their niches, they will automatically acquire higher fitness and secure their survival that way.

Due to the Tuning Phase, by the time Selection Stage starts, each individual presents its topology in roughly the best light it can reach within reasonable time. This is due to the consistent application of Parametric RIM to each DXNN during targeted tuning, and that only after a substantial number of continues failures to improve is the individual considered to be somewhere at the limits of its potential. Thus each NN can be judged purely by its fitness rather than have a need for artificial protection. When individuals are artificially protected within the population, more and more Neurons are added to the NN unnecessarily, thus producing the dreaded topological bloat. Topological bloating dramatically and catastrophically hinders any further improvements due to a greater number of Neurons unnecessarily being in the NN and needing to have their parameters set *concurrently* to just the right values to get the whole system functional. An example of such topological bloating was demonstrated in the robot arm control experiment using NEAT and EANT2 [7]. In that experiment, NEAT continued to fail due to significant neural bloating, whereas EANT2 was successful. Once the NN bloats past a certain size, it simply can not find a solution due to the high number of Neurons that need to have their parameters set concurrently to a proper value. At the same time, most TWEANN algorithms allow for only a small number of perturbations to be applied at any one instance. Once a NN passes some topological bloating point, it simply can not generate enough of concurrent perturbations to fix the faulty parameters of all the new neurons it acquired. In DXNN, through the use of Targeted Tuning and RIMs applied during the Tuning and Topological Mutation phases, we can successfully avoid bloating. Indeed, as will be demonstrated during the experiments in later sections, the DXNN platform consistently produces highly compact NN solutions.

### 4.4 The "Competition" Selection algorithm:

When all NNs have been given their fitness rating, we must use some method to choose those NNs that will be used for offspring creation. DXNN platform uses a selection algorithm I call "Competition", which tries to take into account not just the fitness of each NN, but also the NN's complexity. This selection algorithm is composed of the following steps:
  1. Calculate the average energy cost of the Neuron using the following steps:
      TotEnergy = DXNN(1)_Fitness + DXNN(2)_Fitness...
      TotNeurons = DXNN(1)_TotNeurons + DXNN(2)_TotNeurons...
      AverageEnergyCost = TotEnergy/TotNeurons

  2. Sort the DXNNs in the population based on fitness. If 2 or more DXNNs have the same fitness, they are then sorted further based on size, more compact solutions are considered of higher fitness than less compact solutions.

  3. Remove the bottom 50% of the population.

  4. Calculate the number of alloted offspring for each DXNN(i):
      AllotedNeurons = (Fitness/AverageEnergyCost),
      AllotedOffsprings(i) = round(AllotedNeurons(i)/DXNN(i)_TotNeurons)

  5. Calculate total number of offspring being produced for the next generation:
      TotalNewOffsprings = AllotedOffsprings(1)+...AllotedOffsprings(n).

  6. Calculate PopulationNormalizer, to keep the population within a certain limit:
      PopulationNormalizer = TotalNewOffsprings/PopulationLimit

7. Calculate the normalized number of offspring alloted to each DXNN:
NormalizedAllotedOffsprings(i)=
round(AllotedOffsprings(i)/PopulationNormalizer(i)).

8. If NormalizedAllotedOffsprings (NAO) == 1, then the DXNN is allowed to survive to the next generation without offspring, if NAO > 1, then the DXNN is allowed to produce (NAO -1) number of mutated copies of itself, if NAO = 0 the DXNN is removed from the population and deleted.

9. The Topological Mutation Phase is initiated, and the mutator program then passes through the database creating the appropriate NAO number of mutated clones of the surviving individuals.

From this algorithm it can be noted that it becomes very difficult for bloated NNs to survive when smaller systems produce better or similar results. Yet when a large NN produces significantly better results justifying its complexity, it can begin to compete and push out the smaller NNs. This selection algorithm takes into account that a NN composed of 2 Neurons is doubling the size of a 1 Neuron NN, and thus should bring with it sizable fitness gains if it wants to produce just as many offspring. On the other hand, a NN of size 101 is only slightly larger than a NN of size 100, and thus should pay only slightly more per offspring.

**4.5 Topological Mutation Phase:**

An offspring of a DXNN is produced by first creating a clone of the parent DXNN, then giving it a new unique Id, and finally applying Mutation Operators to it. The Mutation Operators (MOs) that operate on the individual's topology are randomly chosen with uniform distribution from the following list:

1. "Add Neuron" to the NN and link it randomly to and from randomly chosen Neurons within the Core, or one of the Sensors/Actuators.

2. "Add Link" (can be recurrent) to or from a Neuron, Sensor, or Actuator.

3. "Splice Neuron" such that that two random Neurons which are connected to each other are disconnected and reconnected through a newly created Neuron.

4. "Change Activation Function" of a random Neuron.

5. "Change Learning Method" of a random Neuron.

6. "Add Bias" connection (all neurons are initially created without bias).

7. "Add Sensor Tag" which connects a currently unused Sensor present in the SensorList to a random Neuron in the NN. This mutation operator is selected with a researcher defined probability of X. In this manner new connections can be made to the newly added or previously unused sensors, thus expanding the sensory system of the NN.

8. "Add Actuator Tag" which connects a currently unused Actuator present in the ActuatorList to a random Neuron in the NN. This mutation operator is selected with a researcher defined probability of Y. In this manner new connections can be made to the newly added or previously unused actuators, thus expanding the types of tools or morphological properties that are available for control by the NN.

The "Add Sensor Tag" and "Add Actuator Tag" can both allow for new links from/to the Sensor and Actuator programs not previously used by the DXNN to become available to the DXNN. In this manner the DXNN can expand its senses and control over new actuators and body parts. This feature becomes especially important when the DXNN platform is applied to the Artificial Life and Robotics experiments where new tools, sensors, and actuators might become available over time. The different sensors can also simply represent various features of a problem, and in this manner the DXNN platform naturally incorporates feature selection capabilities.

The total number of Mutation Operators (MOs) applied to each offspring of the DXNN is a value randomly chosen between 1 and square root of the total number of Neurons in the parent DXNN. In this way, once again a type of random intensity mutation (RIM) approach is utilized. Some mutant clones will only slightly differ from their DXNN parent, while others might have a very large number of MOs applied to them, and thus differ drastically. This gives the offspring a chance to jump out of large local maxima that would otherwise prove impossible if a constant number of mutational operators were to have been applied every time independent of the parent NN's complexity/size. As the complexity and size of each DXNN increases, each new topological mutation plays a smaller and smaller part in changing the network's behavior, thus a larger and larger number of mutations needs to be applied to produce significant differences to the processing capabilities of that individual. For example, when the size of the NN is a single neuron, adding another one has a large impact on the processing capabilities of that NN. On the other hand, when the original size is a million neurons, adding the same single neuron to the network might not produce the same amount of change in the computational capabilities of that system. Increasing the number of MOs applied when the size of the parent DXNN increases allows us to make the mutation intensity significant enough to allow the mutant offspring to continue producing innovations in its behavior when compared to the parent. At the same time, due to RIM, some offspring will only acquire a few MOs and differ topologically only slightly and thus have a chance to tune and explore the local topological areas on the topological fitness landscape, while others will explore far and wide.

Because the sensors and actuators are represented by simple lists of existing sensor and actuator programs, the DXNN platform allows for the individuals within the population to expand their affecting and sensing capabilities. Such abilities integrated naturally into the NN lets individuals gather new abilities and control over functions as they evolve. For example, originally a population of very simple individuals with only distance sensors is created. At some point a fit NN will create a mutant offspring to whom the "Add Sensor Tag" or "Add Actuator

Tag" mutational operator is applied. When either of these mutational operators is randomly applied to one of the offspring of the DXNN, that offspring then has a chance of randomly linking from or to this new Sensor or Actuator. In this manner the offspring can acquire color, sonar or other types of sensors present in the sensor list, or acquire control of a new body part, some actuator, and further expand its own morphology. These types of expansions and experiments can be undertaken in the artificial life/robotics simulation environments like the Player/Stage/Gazebo Project[9]. Player/Stage/Gazebo in particular has a list of existing sensor and actuator types, making such experiments accessible at a very low cost.

**Fig. 6** The different stages in the DXNN's learning algorithm: Initialization Stage, Tuning Phase, Selection Stage, Topological Mutation Phase.

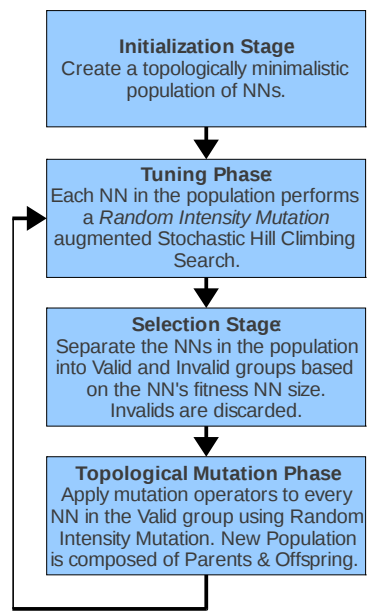

Once all the offspring are generated, they and their parents once more enter the tuning phase to continue the cycle as diagrammed in Fig6.

## 5. Standard Benchmarks:

In this section DXNN will be tested using three standard benchmarks on which other TWEANNs have been tested and thus provide a point of comparison. The first experiment will test whether the DXNN platform can evolve the topology needed to solve the XOR problem when started with a single Neuron without a bias. The second and third experiment will be that of the double pole balancing with and without velocities as specified in [4]. The results produced by DXNN platform will then be compared with other state of the art TWEANNs. In each of the following experiments DXNN platform performs 100 runs with a population size limited to 10. Though DXNN benefits from using large populations, it can manage with very small populations due to its tuning phase and its ability to produce high population diversity due to its topological mutation phase, as will be demonstrated and discussed in section 6. The parameter BaseMaxMistakes was set to 10 in the Double Pole Balancing (DPB) With Velocity Information benchmark, and 20 in the DPB Without Velocity Information benchmark. To make the system comparable to other TWEANNs, only the hyperbolic tangent activation function was used, with the Learning Method parameter restricted to: "none".

## 5.1 XOR Simulation:

The minimal requirement for a TWEANN is the ability to solve the XOR benchmark starting with a single Neuron. To learn to mimic XOR it is necessary for the NN to evolve at least a single hidden Neuron, thus demonstrating DXNN platform's ability to perform topological evolution.

The DXNN platform started with single Neuron topologies without bias. During 100 simulations the platform was able to find the solution 100% of the time, with NN solutions containing 2-3 Neurons. After having demonstrated that it could evolve rudimentary topologies, the DXNN Platform was applied to the double pole balancing problems.

## 5.2. Double Pole Balancing Experimental Setup

The simulation is created using a realistic physical model incorporating friction though fourth order Runge-Kutta integration. A step size of 0.01s was used, with DXNN producing Force values at 0.02s time steps.

The state variables for the problem were as follows:
1. Cart Position
2. Cart Velocity
3. Pole1 Position
4. Pole1 Velocity
5. Pole2 Position
6. Pole2 Velocity

At every time step DXNN receives scaled state variables from the simulation and outputs vector [N], where N is force. N is further scaled to be within the range of -10 and 10.

To pass the test, DXNN must balance 2 poles of different sizes (1m and 0.1m) for 100k time-steps (30 minutes of simulated time). The two poles have initial positions of 4 degrees for the long pole and 0 degrees for the short pole. Both poles must be kept within 36 degrees of the vertical. Furthermore, the cart starts at the center of a 4.8 meter track at X = 0, and must remain within -2.4 and 2.4 meters of that center throughout the experiment. Finally, the Force produced by the DXNN is set to be no less than (1/256)*10N as in [4]. The general setup of the experiment is graphically demonstrated in Fig7.

A single experiment will run until either a maximum number of evaluations is reached, which I shall set to be 50000, or until the problem is solved. The average number of evaluations during the 100 experiments will be compared to the average number of evaluations taken by other TWEANN systems. An evaluation is counted every time DXNN is given a fitness, in other words, each perturbation of weights in the Tunning Phase and subsequent application to the domain counts as an evaluation.

**Fig. 7** DXNN gathers data from the double pole simulation and outputs a force value that is then scaled to be within -10N and 10N, and applied to the cart

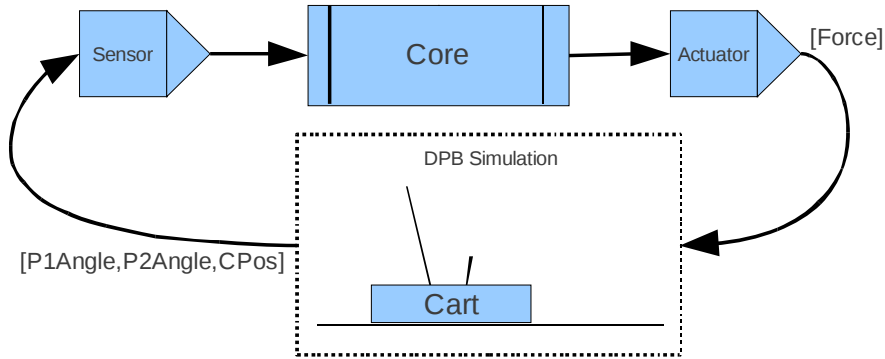

### 5.3. Double Pole Balancing with velocities:

Both, the Double Pole Balancing (DPB) Simulation and the fitness function used are made to the specifications of [4]. The data for Table-1 is taken from [Table-3] in [4], where the average number of evaluations for each system was calculated from 50 runs.

**Table-1** A list of various GA and TWEANN methods, and the # of evaluations they took on average to solve the Double Pole Balancing benchmark with pole and cart velocity information.

| Method | Evaluations |
| --- | --- |
| RWG | 474329 |
| EP | 307200 |
| CNE | 22100 |
| SANE | 12600 |
| Q-MLP | 10582 |
| NEAT | 3600 |
| ESP | 3800 |
| CoSyNE | 954 |
| CMA-EX | 895 |
| DXNN | 725 |

As can be noted by the results, DXNN platform outperforms all other systems, even those that did not have to evolve a topology. The DXNN sizes ranged from 1-2 Neurons, highly compact.

### 5.4 Double Pole Balancing without velocities:

The DPB without velocities is a significantly more complex problem, requiring a recurrent NN to be evolved. Both, the Pole Balancing Simulation and the fitness function used are made to the specifications of [4]. The data for Table-2 is taken from [Table-4] in [4] where the average number of evaluations for each system was calculated from 50 runs.

**Table-2** A list of various GA and TWEANN methods, and the # of evaluations they took on average to solve the Double Pole Balancing benchmark without pole and cart velocity information. Two variations of the problem are benchmarked, one with damping and one without.

| Method | Without-Damping | With-Damping |
|--------|-----------------|--------------|
| RWG    | 415209          | 1232296      |
| SANE   | 262700          | 451612       |
| CNE    | 76906           | 87623        |
| ESP    | 7374            | 26342        |
| NEAT   |                 | 6929         |
| CMA-ES | 3521            | 6061         |
| CoSyNE | 1249            | 3416         |
| DXNN   | 2359            | 2313         |

Using the undamped fitness function the DXNN platform produced highly competitive solutions with DXNN sizes ranging between 2 - 3 Neurons. DXNN Platform lost in its evaluation count only to CoSyNE, which did not have to evolve a topology. When damped fitness function was implemented, the DXNN sizes stayed between 2 - 3 neurons, and outperformed its competitors. The results show that DXNN platform in some cases outperforms even the topologically static methods which do not produce the most compact solutions and can not be applied to the very complex or dynamic problem domains; while at the same time DXNN consistently produced minimal topologies and outperformed all other TWEANNs on this standard benchmark of evolving neurocontrollers.

Based on these results the DXNN Platform is shown to produce results faster than other topology evolving algorithms. The topological compactness can be further increased by increasing the BaseMaxMistakes parameter. By setting this parameter to 100, the DXNN Platform produces NN solutions composed of 2 neurons almost exclusively.

Having demonstrated the system's competence in evolving neurocontrollers, we next analyze the noted high diversity profile of the populations DXNN produces.

## 6. DXNN Population Diversity Profile

In this section I demonstrate that DXNN platform is able to produce populations of excellent diversity. This property is due to the two phase approach. After every tunning phase, the remaining 50% of the population produce topological mutants, which by their very definition have to differ topologically from their parents. DXNN is able to do this efficiently because the tunning phase allows it to thoroughly test every topology in its parameter space. The result is a population that thoroughly explores topological space due to the Topological RIM, while at the same time has its NNs explore and exploit the parameter space due to the Parametric RIM of the tunning phase.

In the following experiments a NN is considered topologically different from the others if it has either a different number of input connections, a different number of output connections, a different number of neurons in total, or a different set of activation functions. The organisms can of course differ even further, since even if two NNs have the same activation function set, the same number of neurons, and input and output links, they can still differ in how the neurons are actually interlinked. But nevertheless, this simple count provides a good method of calculating the lower bounds on diversity in the NN population.

To get the diversity value, I first calculate the noted features for each NN in a population, and then group the NNs based on those features. The total number of different groups is the diversity of the population since each group represents a topologically different set of organisms, a different specie. This diversity value is calculated after the entire population finishes with its tunning phase. In the following graphs, the Generation value is plotted on the X axis while the topological Minimum Diversity value is on the Y axis.

The "Minimum Diversity" value is the number of topologically different organisms in the population during that particular generation. Thus for example in Fig8, when the population size was set to 50 with MaxMistakes set to 10, by the 5$^{th}$ generation nearly 25 different topologies/species existed in the population. In Fig8 the NNs had access only to the Sigmoid Activation Function, while in Fig9 Sigmoid, Sin, and Gaussian Activation Functions were available, in which case the population diversity increased even faster with each generation.

**Fig. 8** DXNNs using only the Sigmoid activation functions. Where "Minimum Diversity" is the number of topologically different organisms in the population. The graph shows that the diversity increases rapidly with each generation, until the majority of organisms in the population is topologically different from one another, no matter what the population size is

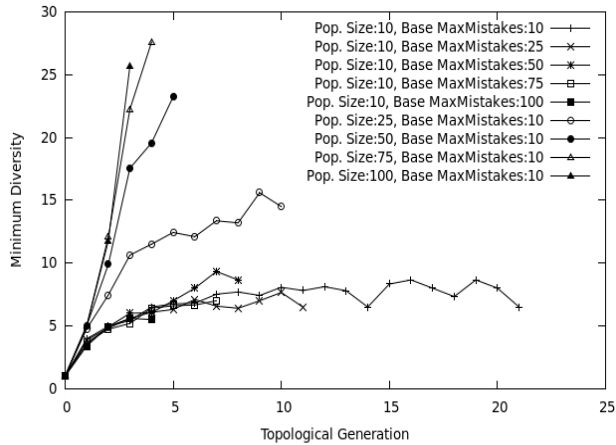

**Fig. 9** The diversity profile of DXNNs populations using Sigmoid, Sin, and Gaussian activation functions. Where "Minimum Diversity" is the number of topologically different organisms in a population. As in Fig8, the population diversity increases with every generation

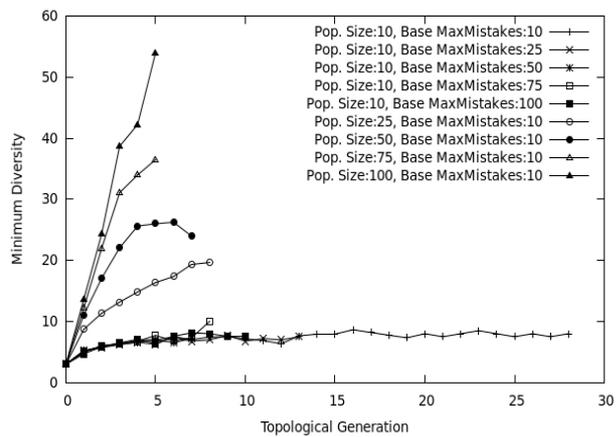

Every point is produced by averaging the diversity values from 50 experiments. The rather extended topological generation range is due to the fact that every set of 50 experiments had 1 or 2 experiments that took more than an average number of generations to complete.

In Fig8 where only the sigmoid activation function is used, we can clearly see that minimum diversity reaches 25-90% (which means that there is roughly a presence of 25-90 different topologies/species in a population of 100) of the population within the first 5 generations. In Fig9 where the available activation function list is composed of sigmoid, sin, and gaussian, the population diversity is even higher. Both of these graphs demonstrate the DXNN system's diversity increases with each generation rather than decreases, until finally the majority of the population is composed different topologies.

This diversity profile signifies that if DXNN Platform is allowed to operate with larger and larger populations, it will explore far and wide within the topological space. Even when the population has found the solution to the problem, we do not see any type of significant drop in the diversity profile. This implies that the population is not being taken over by any one particular topology. This trait is present because 1. Due to the "Competition" selection algorithm, an organism needs orders of magnitude higher fitness to push out all other organisms from the population. And 2. Even if an organism does somehow manage to push out all other organism from the population, the next generation will still be composed of a highly diverse population since the offspring themselves are *topological mutants*.

## 7. Parameter Variation and Oblation

In this section I will be again performing the Double Pole Balancing-With Damping benchmarks. In these benchmarks I will vary the numerous parameters of the DXNN Platform to demonstrate how each one of them affects the system's performance with regards to the total number of evaluations and NN size. In all the following experiments the initial/seed population was set to 10.

### 7.1 The effects of "BaseMaxMistakes" parameter:

The compactness of the solution can be improved by increasing the BaseMaxMistakes parameter. As this variable increases, the DXNN Platform more thoroughly evaluates the true fitness of each topology. But this is done at the expense of the total number of evaluations it takes to reach a solution.

As can be noted from Table-3, though it takes more evaluations to produce a solution as one increases this variable, the average number of neurons in the solution decreases. On the other hand, if BaseMaxMistakes and the Population Size variables are both set to very low numbers, below 10, the system begins to get stuck at times. The system is considered to have failed if it did not solve the problem after 50000 evaluations or 100 Topological Mutation Generations, whichever is reached first.

**Table-3** The variable "BaseMaxMistakes" is changed, affecting the system's performance. The result of increasing the BaseMaxMistakes parameter is the decrease of the average number of neurons in the evolved solutions.

| BaseMaxMistakes | Population_Size | Avg # of Evaluations | Avg # of Neurons |
|---|---|---|---|
| 1 *46% Failure | 10 | 1596 | 8.5 |
| 5 *10% Failure | 10 | 1545 | 4.7 |
| 10 | 10 | 2084 | 3.59 |
| 20 | 10 | 2313 | 2.74 |
| 30 | 10 | 2803 | 2.73 |
| 50 | 10 | 2951 | 2.62 |
| 100 | 10 | 3919 | 2.44 |

## 7.2 The effects of Population_Size

As can be noted in Table-4, DXNN solutions become more compact as one increases the Population Size. Once again we can note that the system begins to fail when the Population Size limit is decreased to very small numbers. This is normal because with such a low populations only 1-2 new organisms are created in each generation, making it possible to get stuck once in a while, unless BaseMaxMistakes is increased to compensate.

**Table-4** In this test the NN population is varied, affecting the average number of evaluations needed and the number of neurons in the resulting solutions.

| Population_Size | BaseMaxMistakes | Avg # of Evaluations | Avg # of Neurons |
|---|---|---|---|
| 5  *4% Failure | 10 | 1968 | 3.74 |
| 10 | 10 | 2084 | 3.59 |
| 20 | 10 | 2214 | 3.1 |
| 30 | 10 | 2274 | 2.79 |
| 50 | 10 | 2953 | 2.61 |
| 100 | 10 | 4576 | 2.38 |
| 5 | 50 | 4215 | 2.76 |
| 10 | 50 | 2951 | 2.62 |
| 20 | 50 | 4526 | 2.34 |
| 30 | 50 | 5805 | 2.35 |
| 50 | 50 | 7538 | 2.29 |
| 100 | 50 | 11881 | 2.13 |

## 7.3 The effects of Weight RIM

One of the claims in this paper is that Parametric RIM is an important part of DXNN, acting as an efficient approach to the exploration and exploitation of the topological and parametric space. In Table-5 I demonstrate what happens when I slowly decrease Parametric RIM intensity. For these experiments, the population limit was set to 10.

**Table-5** In this test Weight RIM is slowly decreased, resulting in system's decreasing performance.

| Weight RIM | BaseMaxMistakes | Avg # of Evaluations | Avg # of Neurons |
|---|---|---|---|
| -Pi to Pi | 50 | 2951 | 2.62 |
| -Pi/2 to Pi/2 | 50 | 3880 | 2.66 |
| -1 to 1 | 50 | 4106 | 2.68 |
| -0.5 to 0.5 | 50 | 5135 | 2.92 |
| -0.3 to 0.3 *22% Failure | 50 | 14594 | 5.94 |

| -0.2 to 0.2 *56% Failure | 50 | 25038 | 11.76 |
| --- | --- | --- | --- |
| -0.1 to 0.1 *100% Failure | 50 | Failed | * |

It can be observed that as the RIM intensity is decreased, the system begins to take longer and longer to solve the double pole balancing problem, and eventually begins to fail completely. At the same time, the average number of evaluations is getting closer to what is seen from the standard evolutionary algorithm systems, 25000+ evaluations. With RIM, even when BaseMaxMistakes is taken to be 50, the DXNN Platform is able to produce solutions with an average number of evaluations in the 3000 area, and NN size of 2-3.

This is as expected, since without RIM the system does not explore the parametric space almost at all. During a crossover two similar topologies would be chosen from which an offspring would be created with a mix of weights from both parents, this would produce a large enough jump in the topological space to give a chance for the solution to jump out of a parametric local maxima and find a better solution. This is the same job that is being accomplished by Parametric & Topological RIM, by providing a chance for the NN to undergo a significant mutation that affects a large number of neurons and weights within each neuron. Thus RIM does indeed take over the exploratory duties that are usually accomplished by the crossover algorithms. But RIM can be more easily and effectively controlled, is simpler to implement, and based on the benchmarks, it performs better too.

It is my opinion that crossover is a vestigial, and inefficient method that somehow was picked up from the field of biology. But we don't need it, technology allows us to create algorithms and methods that are much more robust and dynamic than any biologically inspired crossover approach. Parametric and Topological RIMs are more efficient, easier to implement, easier to experiment with, easier to vary and control, and as the benchmarks demonstrate and will further demonstrate in the following ALife simulations, is an excellent and perhaps superior alternative.

## 8. Open-ended "Flatland" ALife experiments

Some of the more complex problems are those where the NN has to evolve solutions in ever changing environments, with multiple sensory inputs and outputs. These problems are within the field of Robotics and Artificial Life. These are no longer just benchmarks or toy problems, but real world applications with results that have significant consequences. These simulations demonstrate whether the system in question can evolve NNs which behave appropriately within the environment they inhabit. In this chapter I will use the DXNN Platform to evolve neurocontrollers of organisms inhabiting increasingly more complex 2d environment scenarios, and demonstrate that this system can evolve highly efficient and indeed optimal behavioral strategies.

I will apply the DXNN platform to ALife food gathering and predator-prey simulations. In these simulations DXNN will be used to evolve the brains of artificial robots inhabiting a two dimensional world, a flatland. The first experiment will demonstrate that the DXNN Platform can rapidly evolve robots which can successfully use range and color based sensors, and 2 wheel based differential drive actuators to move around the 2d environment and gather food. The second experiment will scatter poison among the food and demonstrate that the DXNN platform can evolve NNs that gather food and avoid poison. Finally, the third experiment will evolve 2 separate populations inhabiting the same 2d environment, a prey population which feeds on plants, and a predator population that feeds on prey. This simulation will demonstrate that DXNN can be very successfully be applied to the problems in the complex field of co-evolution, with very impressive results. Recorded videos of the simulations at different stages of evolution are available as supplementary material to this paper.

**8.1 General simulation setup parameters:**

These ALife simulations are conducted in a 2 dimensional environment without walls. The simulated organisms are based on Khepera robots. The simulated robots are represented as circles, using a differential drive to move around the environment. The environment has a fully implemented collision detection, and all robots can push each other. When two robots are pushing each other, the one that has more energy is the one whose push is realized. The SensorList available to the organisms contains a Range and a Color Sensor. The output of a range sensor is a vector composed of range values. The range values are calculated through ray casting. The color sensor also uses ray casting, but instead of producing lists of ranges, it produces a list of floating point encoded colors. Each sensor casts rays within a 90 degree arc of where the robot is pointing, and the number of rays into which the 90 degree arc is divided is determined by the resolution of that sensor, 5 in these simulations. The simulated robots, sensors, and color encoding is presented in Fig10.

**Fig. 10** The 2d representations of Prey, Predator, Plant, Poison, and the color encoding scheme

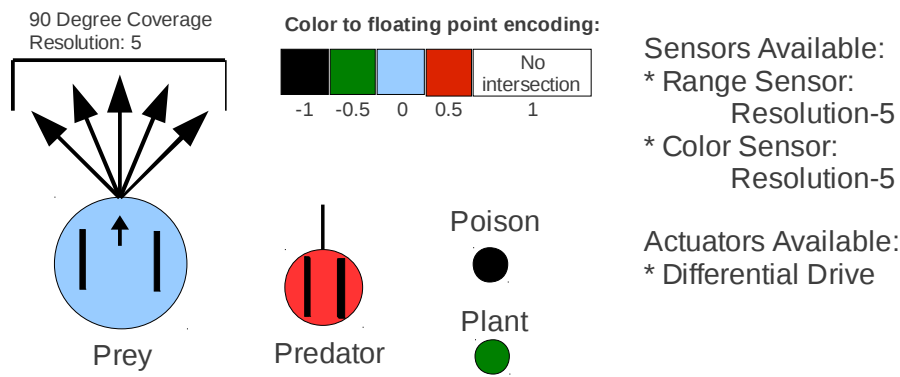

All simulated robots start with 1000 energy points. Every action the simulated robot makes costs energy. Movement energy cost is 1 energy point per 1m/s. The turning energy cost is 1 energy point per 90 degrees/s. When the robot moves at the fraction of the speed or turns a fraction of 90 degrees per step, the energy used for that action is linearly scaled to the same fraction. Another 0.1 energy points is subtracted from the robot during the use of the differential drive actuator with any level of intensity, ensuring that even if the robot sits still, it will still eventually lose all of its energy and be removed from the environment. A robot's age increases by 1 age point every time it uses its differential drive actuator. The robot dies of old age after reaching 20000 age points, if it does not die from other causes (losing all its energy, or being eaten by a predator) before that. This will prevent robots from living indefinitely, and should produce an evolutionary push towards a strategy that maximizes the total number of prey caught or plants eaten as quickly as possible.

    The fitness given to a simulated robot when it dies is calculated as follows:
        AlivePoints = StepsLived/1000 for the first 1000 steps, and 0 for the rest.
        *Prey*: Fitness = AlivePoints + PlantsEaten
        *Predator*: Fitness = AlivePoints + PreyEaten
Where the term Step stands for the single use of the differential drive actuator.

    The ALife simulation uses an "Augmented Competition" (AC) selection algorithm. The AC selection algorithm keeps a list of size "PopulationSize" of dead NN genotypes, this list is called the "dead pool". When a NN dies its genotype and fitness is entered into this list. If after entering the new genotype into the dead pool the list's size becomes greater than PopulationSize, then the lowest scoring DXNN genotype in the dead pool is removed. In this manner the dead pool is always composed of the top performing "Population Size" number of ancestor genotypes. In this augmented version of the selection algorithm, the AllotedOffspring variables are converted into normalized probabilities used to select a parent from the dead pool to produce a mutated offspring. Finally, there is a 10% chance that instead of creating an offspring, the parent itself will enter the environment. Using this "re-entry" system, if the environment or the manner in which the fitness is alloted changes, the old strategies and their high fitness scores can be re-evaluated in the changed environment to see if they deserve to stay in the dead pool, and if so, what their new fitness should be. This selection algorithm also has the side effect of having the dead pool implicitly track content drift of the problem it's used on.

    Due to the computational cost of these experiments, each version of the simulation is ran 10 times, for 25000 evaluations/simulation in the case of simple Food Gathering, 50000 evaluations/simulation in the case of Food Gathering with poison, and 100000 evaluations/simulation in the Predator Vs. Prey experiment. An evaluation is counted every time a fitness of an organism is calculated. The Activation Function list contained: tanh, sin, linear, gauss, sqrt, absolute value, and log. BaseMaxMistakes variable was set to 20. Finally, for all 3 of the following experiments, "Add Sensor Tag" and "Add Actuator Tag" mutation operators will have a probability of 10% of being chosen by a NN during its topological mutation phase.

## 8.2 Simple food gathering:

In this simulation the environment is populated by Prey and Plants. The Prey inhabiting the 2d environment are controlled by NNs. The Prey are colored blue and the plants green. Each plant contains 500 energy points, which are transferred to the Prey when the plant is eaten. To eat the plant the simulated robot needs to run over it. Four different versions of these simulations are performed:

1. The initial Prey population organisms are generated containing both the Range and the Color sensor, in which case the NNs only need to learn how to use these sensors and how to move around and eat plants. Population Size is set to 10.
2. The initial Prey population organisms are generated connected only to the Range sensors. Since color plays a pivotal role in discriminating between other prey and the plants, this simulation requires that each Prey organism evolves a connection to the Color sensor during its evolution. Population Size is set to 10.
3. Same as 1, but population size is set to 20.
4. Same as 2, but population size is set to 20.

Each simulation version is performed 10 times, with each simulation running for 25000 evaluations.

### 8.2.1 Results

Fig11 presents a graph of the Avg. Fitness Vs. Evaluations, Fig12 presents a graph of Avg. # of Neurons/Organism Vs. Evaluations, and Fig13 presents a graph of Avg. Diversity Vs. Evaluations. These graphs were done by calculating averages between the 10 simulations for each version of the problem, every 500 evaluations.

**Fig. 11** Simple food gathering: Avg. Fitness vs. Evaluations. The P-10 and P-20 "Range and Color" simulations initially started with Range and Color sensors, with population sizes of 10 and 20 respectively. The P-10 and P-20 "Range" simulations initially started with only the Range sensors, and had to evolve connections to the Color sensors during the simulation runs.

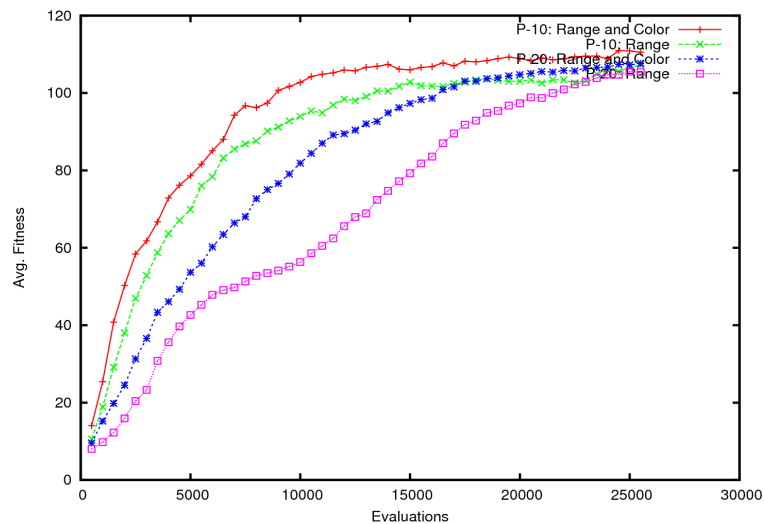

**Fig. 12** Simple food gathering: Avg. # of Neurons/Organism vs. Evaluations

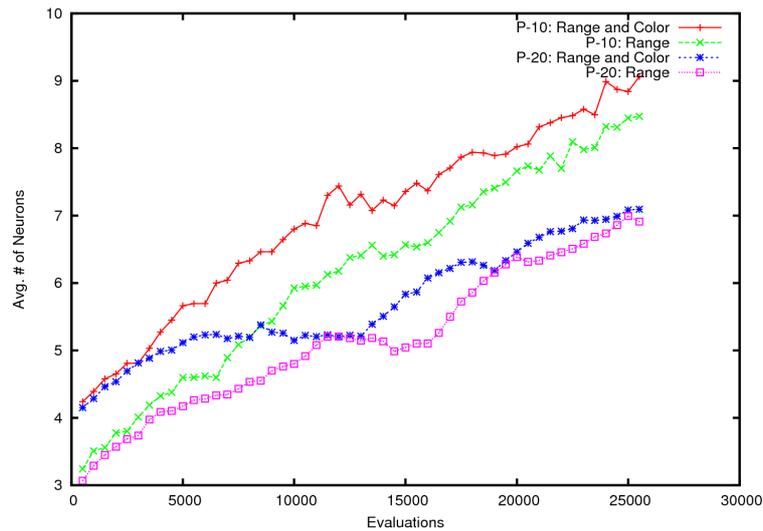

**Fig. 13** Simple food gathering: Avg. Diversity vs. Evaluations

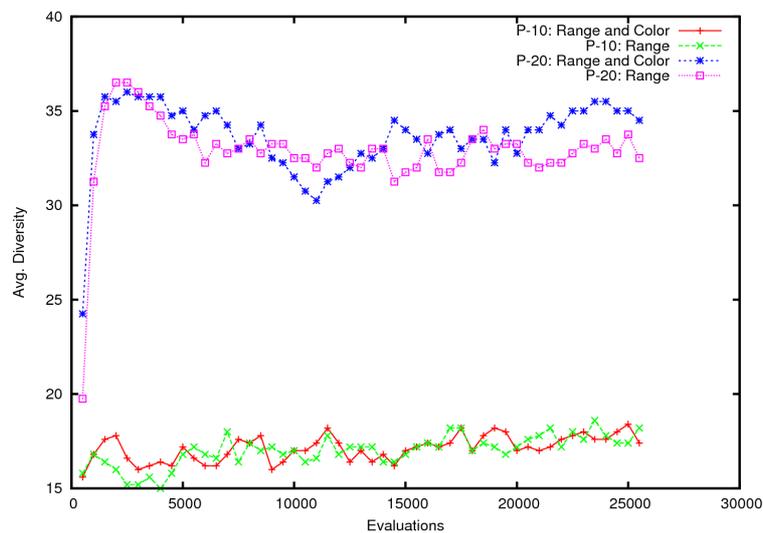

*8.2.2 Discussion*

From the graphs we can see that the prey quickly learn to navigate and move towards the food sources. The prey begin to demonstrate adaptation after only 5000 evaluations. Initially the organisms swarmed the nearest plants, and since only a single robot can eat the plant after which the plant disappears and is respawned in another location, the other robots only waste their energy when swarming strategy is used. In every simulation after about 10000 evaluations a new type of behavior evolves, the robots take into account not only the plants, but also other robots, and as we can see in (Online Resource 1) that rather than moving in swarms, each robot attempts to go for a plant that is not being swarmed by other robots. This final strategy is the most effective, and it is retained until the end of the simulation. The average NN sizes of the organisms towards the end of the simulations ranged from 6-10 neurons, very compact.

It is interesting to note that the simulations where the NNs started with only range sensors did rapidly evolve connections to the color sensors, and then learned to use them to reach the fitness score close to the organisms which started initially with both the range and color sensors. Finally, the population diversity, as shown in Fig13, stabilized and never dipped even when very successful strategies were discovered, thus demonstrating that DXNN does not suffer from population diversity drops.

Population diversity was calculated using the "Total Population Set", composed of the deadpool organisms and the organisms active in the environment. In the population size 20 version (Total Population Set = 40, 20 in the deadpool plus 20 active) of the experiment, diversity ranged from 30 to 38. In the population size 10 version (Total Population Set = 20, 10 in the deadpool plus 10 active) of the experiment, diversity ranged from 14 to 18. Thus the population always maintained a highly diverse set of topologies.

There were no significant differences in the evolved adaptation behaviors in the 4 versions of these simulations. A number of simulations in all 4 versions were able to generate organisms that reached ~120 fitness score.

## 8.3 Dangerous food gathering

In this simulation the 2d environment is filled with prey controlled by the NNs, plants as in the previous experiment, but also poisonous plants that are scattered around in the same area as the plants. The prey are colored blue, the plants green, and the poisonous plants black. Each plant contains 500 energy points, and each poisonous plant contains -2000 points. Similarly to 9.2, four total versions of the simulation are performed. Each simulation version is performed 10 times, with the simulations running for 50000 evaluations.

### 8.3.1 Results

Fig14 presents a graph of the Avg. Fitness Vs. Evaluations, Fig15 presents a graph of Avg. # of Neurons/Organism Vs. Evaluations, and Fig16 presents a graph of Avg. Population Diversity Vs. Evaluations.

**Fig. 14** Dangerous food gathering: Avg. Fitness Vs. Evaluations

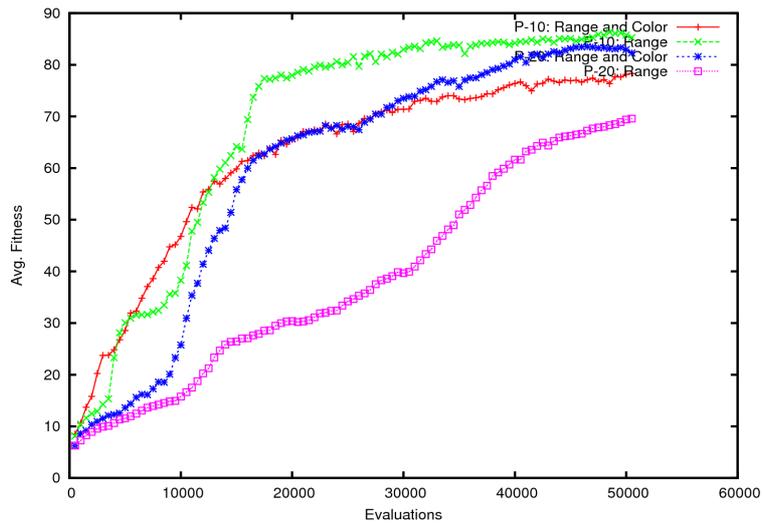

**Fig. 15** Dangerous food gathering: Avg. # of Neurons/Organism Vs. Evaluations

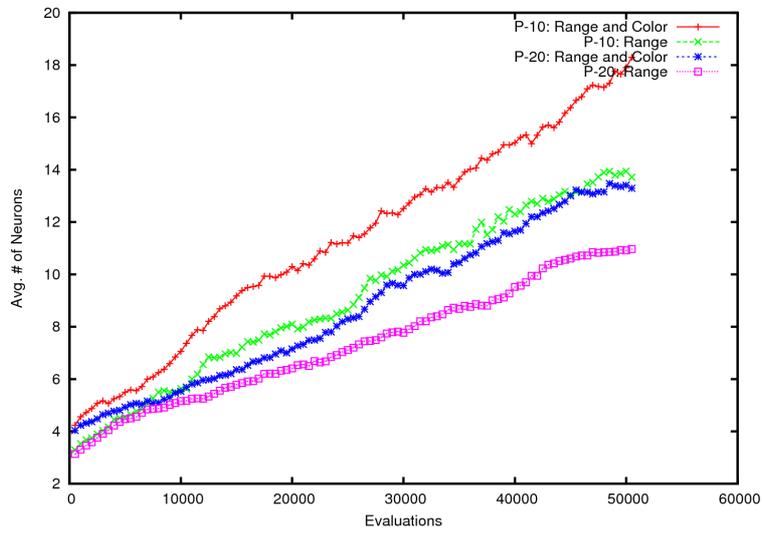

**Fig. 16** Dangerous food gathering: Avg. Diversity Vs. Evaluations

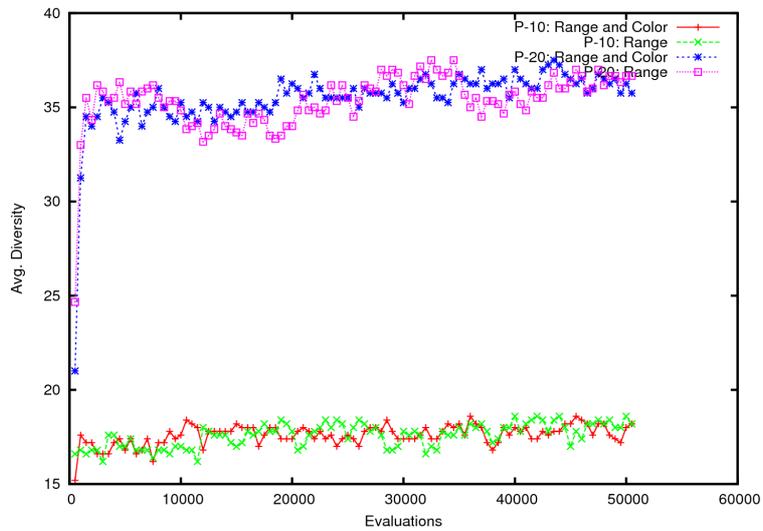

*8.3.2 Discussion*

As the "Fitness Vs. Evaluations" plot demonstrates, the prey eventually learn to pick up the plants even with the presence of poison. On average, just after 10000 evaluations the Prey evolved the ability move around to gather food and discriminate it against poison. After 40000 evaluations, in most experiments the Prey reached a stable fitness score. The strategy that evolved on most of the occasions is to gather as many plants as possible as quickly as possible, even if it meant consuming poisonous plants every once in a while, as shown in the recorded video in (Online Resource 2). The average NN sizes ranged from 10 to 20 neurons. The populations where organisms initially started with only the Range sensors, evolved more compact topologies, but reached the fitness similar to the "Range and Color" populations slower.

In about 25% of the experiments, the organisms evolved a strategy which followed a very conservative approach, where avoiding poison took precedence over gathering food. In those experiments the average fitness score was ~60. In the rest of the simulations, the highest achieved fitnesses was ~120. A number of simulations in all 4 versions were able to generate organisms that reached ~110 fitness score.

In general, the fact that navigation through the poison infested environment is at all possible with such low resolution (R=5) sensors is already remarkable, for it means that the Prey learned to move around and discern what is poison and what is plant even when the two are in very close proximity. In this experiment DXNN is again able to very rapidly evolve appropriate behaviors, even in the cases when it's required to acquire connections to the Color Sensor during its evolutionary path. Also as demonstrated in Fig16, diversity and topological exploration did not waver when behavioral strategies began to stabilize.

### 8.4 Predators Vs. Prey

The predator robot, represented by a red circle with an arrow, can not eat plants and can only gain energy by eating prey robots. When a predator comes in contact with a prey, the prey dies and the predator's energy is incremented by the total amount of energy that the prey had at that point in time. The prey, represented by a blue circle, can gain energy only by eating plants. Plants are represented by small green circles. After being eaten, the plant regrows at a random coordinate calculated by taking a random number from a uniform distribution between 0 to 800 for X, and 0 to 500 for Y. A plant gives the Prey 500 energy points as in the previous 2 experiments. An organism dies when its energy level drops to zero.

There is a slight difference in the location where Plant/Prey and Predators are spawned and re-spawned. As noted above, the Plants are spawned in the $0 < X < 800$ and $0 < Y < 500$ area. The Prey are also spawned in the same region. On the other hand the Predators are spawned and re-spawned in the $800 < X < 1400$ and $0 < Y < 500$ region. This is done to ensure that there are no instances where Predators and Prey spawn in the same location, resulting in the Prey being killed too quickly without any type of evaluation of its strategy.

Due to the high variability of the dynamic Predator/Prey behavior, and evolutionary paths which were highly dependent on which of the species learned to exploit the environment first, no useful graphs of Avg. Fitness vs. Evaluations were made. When Prey learned to move around and eat plants early, the Predators had harder time evolving strategies hunting the already adapted and mobile prey. When the Predators learned to hunt Prey early, the prey had harder time evolving food gathering strategies, since a lot of the prey would get eaten before their strategies were evaluated for fitness. But though the evolutionary paths varied wildly and calculated performance averages were not useful, during every simulation both of the species eventually learned to exploit and adapt within the environment, by eating plants or hunting for prey respectively.

*8.4.1 Discussion*

The behaviors that evolved are most interesting. Initially the prey learned to simply approach and eat the plants, and the predators when near a prey learned to approach and eat the prey. But the behavior evolved to become more advanced. On average after ~70000 evaluations in most experiments the predators learned how to sort of push and hide behind plants, and then push the plants around in front of them until a prey approached, at which point the predator quickly attacks and tries to catch the prey, as shown in (Online Resource 3). This type of clever ambushing feels remarkably organic in its nature. Since the prey have to eat plants to increase their chances of creating offspring, they have no choice but to sooner or later to try and run for the plant. Nevertheless, the prey also eventually learned to be more cautious and evasive, as shown in (Online Resource 4).

As mentioned, another interesting observation was that during all the experiments, whenever the Prey were the first to learn how to gather plants, the Predators took a lot longer to catch up and learn how to hunt the Prey. When the Predators learned how to catch the Prey first, the Prey took longer to learn how to gather food. This makes sense, because when Prey learn how to gather food they become much harder targets due to their constant movement, and so evolving a hunting strategy in such an already dynamic environment becomes more difficult for the predators. On the other hand, when the Predators learn how to catch prey early in the experiment, the Prey do not have enough time to learn to gather plants, since they get eaten before their plant eating strategy is properly evaluated. The evolutionary path of both organisms was very sensitive to such early conditions. Eventually though, both species of the organisms reached a level of proficiency in food gathering and hunting. Similar behavioral strategies eventually emerged in both, the experiments whose populations were seeded with NNs which initially only contained the Range sensors, and those containing Range and Color sensors from the start.

**8.5 Discussion**

DXNN produced the strategies for all 3 problems very rapidly. The strategies were efficient, and in the case of Predator Vs. Prey, surprisingly organic. The system also demonstrated its ability to consistently maintain population diversity throughout the experiments, and the ability to successfully incorporate new sensory organs during the evolutionary process. Thus I believe that DXNN proved itself to be very capable in the co-evolutionary and ALife domain, which gives hope for its performance in future applications to robotics.

## 9. Summary and Conclusion

In this paper I presented a novel TWEANN called DXNN, a memetic algorithm based Topology and Weight Evolving Artificial Neural Network system that separates the parametric and topological mutation phases. DXNN uses a database friendly, tuple based, and human readable genotype encoding approach. Other features of the system include: Targeted Tuning during which stochastic hill climbing is applied only to the newly created or augmented topological elements of the NN, Random Intensity Mutation algorithm which allows for DXNN to both, produce small mutations aimed at fitness landscape exploitation, and large scale mutations aimed at exploration, and replacing the crossover algorithms by acting as a more controllable version of the same. Finally, due to the clear separation of parametric and topological mutation phases, the system is able to have a diverging population diversity profile, as was demonstrated in section 6 and the artificial life experiments of section 8.

DXNN demonstrated its excellent performance when evolving controllers in the standard double pole balancing benchmarks when compared to other state of the art TWEANNs. It further demonstrated its ability to evolve effective behaviors in the ALife simulations. I believe that these ALife simulations prove that the system would also work just as well in the field of robotics. To test this, the next step is the change from the "Flatland" 2d ALife environment used in this paper, to the Player/Stage/Gazebo[9] robot simulator which allows for a seamless transition from simulated environments and robots, to real ones. Based on the standard DPB benchmarks and evolutionary effectiveness in ALife simulations, DXNN proved itself to be a strong and effective TWEANN.